\documentclass[review]{elsarticle}

\pdfoutput=1
\journal{Journal of Neurocomputing}

\RequirePackage{fix-cm}
\usepackage{graphicx}
\usepackage{subfigure}
\usepackage{algorithm}
\usepackage{setspace}
\usepackage{amsmath}
\usepackage{bm}
\usepackage{amssymb}

\usepackage[mathscr]{eucal}
\usepackage{algorithmicx}
\usepackage{algpseudocode}
\usepackage{booktabs}
\usepackage{longtable}
\usepackage[marginal]{footmisc}
\usepackage{float}
\usepackage{xcolor}
\usepackage{wrapfig}

\usepackage{color}

\begin{document}

\begin{frontmatter}

\title{Leveraging Multi-level Dependency of Relational Sequences for Social Spammer Detection
%\tnoteref{mytitlenote}
}
%\tnotetext[mytitlenote]{Fully documented templates are available in the elsarticle package on \href{http://www.ctan.org/tex-archive/macros/latex/contrib/elsarticle}{CTAN}.}

%% Group authors per affiliation:
%\author{Elsevier\fnref{myfootnote}}
%\address{Radarweg 29, Amsterdam}
%\fntext[myfootnote]{Since 1880.}

%% or include affiliations in footnotes:
\author[mymainaddress]{Jun Yin}
\ead{Jun.Yin-2@student.uts.edu.au}
\author[mymainaddress]{Qian Li}
\ead{Qian.Li@uts.edu.au}
\author[mymainaddress]{Shaowu Liu}
\ead{Shaowu.Liu@uts.edu.au}
\author[mysecondaryaddress]{Zhiang Wu}
\ead{zawuster@gmail.com}
\author[mymainaddress]{Guandong Xu\corref{mycorrespondingauthor}}
\cortext[mycorrespondingauthor]{Corresponding author}
\ead{Guandong.Xu@uts.edu.au}

%\author[mymainaddress,mysecondaryaddress]{Elsevier Inc}
%\ead[url]{www.elsevier.com}

%\author[mysecondaryaddress]{Global Customer Service\corref{mycorrespondingauthor}}
%\cortext[mycorrespondingauthor]{Corresponding author}
%\ead{support@elsevier.com}

\address[mymainaddress]{Advanced Analytics Institute, School of Computer Science,\\ University of Technology Sydney, Australia}
\address[mysecondaryaddress]{School of Information Engineering,\\ Nanjing Audit University, Nanjing, China}

\begin{abstract}
Much recent research has shed light on the development of the relation-dependent but content-independent framework for social spammer detection. This is largely because the relation among users is difficult to be altered when spammers attempt to conceal their malicious intents. Our study investigates the spammer detection problem in the context of multi-relation social networks, and makes an attempt to fully exploit the sequences of heterogeneous relations for enhancing the detection accuracy. Specifically, we present the Multi-level Dependency Model (\emph{MDM}).
The~\emph{MDM} is able to exploit user's long-term dependency hidden in their relational sequences along with short-term dependency. 
Moreover, \emph{MDM} fully considers short-term relational sequences from the perspectives of individual-level and union-level, due to the fact that the type of short-term sequences is multi-folds.
Experimental results on a real-world multi-relational social network demonstrate the effectiveness of our proposed \emph{MDM} on multi-relational social spammer detection.\end{abstract}

\begin{keyword}
\texttt Social Spammer\sep Relational Sequence, Multi-level Dependency Embedding  \sep Classification
\end{keyword}

\end{frontmatter}

%\linenumbers

\section{Introduction}
\label{intro}
Social network is a space where all people interact with each other 
and anyone can read, publish, and share content. 
While social network has several groundbreaking benefits, 
it is also a breeding ground for social spammers. 
Everyone can reach thousands of people on social network instantaneously, their behaviours yet are shielded by anonymity.
Consequently, extensive misbehaviours such as the dissemination of fraudulent information and false comments may occur. 
Spamming behaviour we discussed in this paper is not limited to a single malicious activity. Instead, users with any kind of malicious activities will be labeled as spammers.
For example, 
when marketers send unwanted advertisements 
or steal user information by pointing users to malicious external pages 
and around $83\%$ of social networks’ users 
have received more than one unwanted friend request 
or message~\cite{stringhini2010detecting}. 
Such a behaviour seriously affects the development of social network, which is required to be detected in advance so as to maintain a healthy social network. 

Considerable efforts have been devoted to transform the spammer detection into a classification problem. 
As spammers are the people who spread misinformation to the public, content-based features~\cite{Jindal07ICDM1,Jindal07WWW,mukherjee2013yelp,li2011learning,lim2010detecting} are considered as the most representative features for the classification-based detection.
For example, 
Grier et al.~\cite{grier2010spam} extract some content-based features according to the analysis on Twitter spam, e.g., the ratio of tweets containing URL, the ratio of tweets with special characters. 

As the user privacy in social network has attracted increasing attention,
the metadata in social network, especially the contents, is relatively scarce. 
Rather than exploiting the explicit content, researchers resort to the network topological structure that is the implicit attribute of social network with multiple \emph{relations} for social spammer detection~\cite{brophy2017collective,chakraborty2016recent,manaskasemsak2019opinion}, where \emph{relations} refer to the interactions between users (e.g., sending messages, viewing profile, thumbs up, forward posts etc.). 
For instance, the network topological graph is generated for each relation~\cite{fire2012strangers,benczur2005spamrank,bhat2014spammer}. Meanwhile, spammers are assumed to be the important nodes with more links from other nodes in the graph, graph-based features are then extracted by using several graph analytic methods (e.g., \emph{Triangle count}~\cite{schank2007algorithmic} and \emph{k-core}~\cite{alvarez2006large}.)

Nevertheless, 
these graph-based methods merely detect the spammers using the single-relation, which however violates the fact that spammers may be connected to normal users in terms of multiple relations.
Hence, 
the chronological sequence of relations (i.e., [viewing profile$\rightarrow$ thumbs up$\rightarrow$ forward posts$\rightarrow$ sending messages]) is usually considered to extract the sequence-based features on multi-relational social network~\cite{fakhraei2015collective,peng2004augmenting,fu2018combating}.
For example, Fakhraei et al.~\cite{fakhraei2015collective} 
define a short sequence segment of $k$ consecutive actions, called a $k$-gram,
and use the number of occurrence a $k$-gram sequence to partly disclose the difference between the spammers and the normal users.
Although $k$-gram features capture the short-term aspects of the sequence, they may miss the long-term dependency of the sequence.
Instead, to capture the salient information from longer sequence chains, and to study the predictive power of this information,
mixture of Markov models are utilized by Peng et al.~\cite{peng2004augmenting} to overcome the limitation of small $k$. 
Specifically, Peng et al. use the ratio of posterior probabilities and their logarithms as a small feature-set, which is identified as a small set of important sequence from long sequence chains, for their classifier.

In general, existing sequence-based methods either exploit the long-term or the short-term dependency, which may be more likely to ignore the underlying correlations between them. Moreover, most existing sequence-based methods trained merely on the limited training datasets tend to be overfitted~\cite{peng2004augmenting,li2014search}.
What if new spammers deliberately do not follow the known behaviour pattern that they usually have?
To address this issue, 
our goal is to expose the deeper information hidden behind the sequence so as to identify their abnormal behaviours accurately.
Inspired by deep sequential networks~\cite{yu2019multi,zhou2018personalized,tang2018personalized},
we exploit both the \emph{long-term} and \emph{short-term} dependencies to fully learn the deeper complementary information underlying users' multi-relational sequences.
Specifically, long-term dependency models the users overall behaviours on multi-relational social network based on their whole day's relational sequences, 
while short-term dependency exploits the information of partial behaviours with the most recent $n~(1\leq n < 10)$ relational sequences. Moreover, we exploit the short-term dependency in terms of individual-level and union-level. On the individual-level, we only consider one relation, user performed recently, that may trigger his/her next behavior. While on the union-level, we capture the collective influence among a union of relations that the user performs.

In this paper,
we propose a novel
\emph{Multi-level Dependency Model} (\emph{MDM}),
which exploits user's behaviours in terms of long-term and short-term dependency from both individual-level and union-level.
The individual-level dependency considers only a single recent behaviour that may trigger subsequent behaviours. In contrast, the union-level dependency considers the collective influence among a union of relations that are involved in the user's short-term behaviour sequence.
\emph{MDM} is capable of exploiting the deeper information hidden behind users' relational sequence and hence improves the performance of multi-relational social network spammer detection.
The main contributions of our paper can be summarized as:
\begin{itemize}
	\item \emph{MDM} is capable of exploiting user's long-term behaviours hidden in their multi-relational sequential behaviours along with short-term relational behaviours from multiple perspectives, which largely overcomes the limitation of one-sided exploration of sequences.
	\item To model the short-term dependency, \emph{MDM} exploits the relational sequences from both individual-level and union-level perspectives. Besides, the residual network in \emph{MDM} can learn high-order sequential dependency among multi-relations.
	\item Extensive experiments on real-world data demonstrate that \emph{MDM} outperforms the state-of-art baselines of spammer detection.	
\end{itemize}

The following sections will be organised as follows.
In Section~\ref{sec:rw}, we discuss the related work
and outline the limitations of the methodologies in the literature.
We formulate the spammer detection problem and illustrate the overall framework of our proposed \emph{MDM} in Section~\ref{sec:preliminary}.
Section~\ref{sec:method} provides the technical details of our~\emph{MDM},
followed by extensive experimental results
in Section~\ref{sec:experiment}.
Finally, we conclude our work and give future plan in Section~\ref{sec:conclusion}.

\section{Related Work}
\label{sec:rw}
In the literature, extensive work have been proposed
to extract features of the spammers in social media,
including e-commerce sites~\cite{hussain2020spam,Fayazi15SIGIR,Wu15ICDM,gong2020attention} and social network sites~\cite{Arjun13KDD,shehnepoor2017netspam,masood2019spammer,wu2020graph}.
Generally, these methods can be categorized into four categories:
\textit{content-based}~\cite{sriram2010short,bakshi2016opinion,liu2010sentiment}, 
\textit{behaviour-based}~\cite{benevenuto2008identifying,parameswaran2010game,lin2013analysis},
\textit{graph-based}~\cite{brophy2017collective,krestel2008using,bhat2013community} 
and \textit{sequence-based} methods.~\cite{fakhraei2015collective,peng2004augmenting,wang2017clickstream}

In early studies of email spams and e-commence spams,
reviews/emails containing similar content have a high probability to be spams~\cite{Jindal07ICDM1,Jindal07WWW}.
Various of content-based features are designed to detect such spams in e-commerce and emails.
While content-based features mostly rely on natural language processing methods, including text classification~\cite{sriram2010short}, text sentiment analysis~\cite{bakshi2016opinion} and text orientation analysis~\cite{liu2010sentiment}.
%, e.g.,
%average length in number of words~\cite{mukherjee2013yelp},
%ratio of objective words~\cite{li2011learning}.

Both the amount of information and the rate of generation in social networks far exceed that of e-commerce sites and emails. In addition, social networks generally have restrictions on the number of words in text, and because of the user privacy protection, content-based dataset is difficult to collect.
Benevenuto et al.~\cite{benevenuto2008identifying} first applied statistics on spammer behaviour in \emph{YouTube}. They manually labeled the dataset to establish training data, and then analyzed the behaviours of the labeled spammers, and defined their characteristics. They used three feature selection algorithms in \emph{Weka} to evaluate the discrimination power of each spammer behaviour feature, and used traditional supervised classification methods to spammer classification.
This method is a representative method in the field of spammer detection in social networks, that is, based on user behaviour characteristics to identify the network spammer. Subsequent studies on spammer detection are inspired by this method, adding features or optimizing detection methods to improve the accuracy of spammer detection in social networks. 

In addition to behaviour-based methods, users in social networks will gradually form a user-centric social circle through interactive behaviours, the social relationships between users often contain rich information. 
Compared with normal users, spammers in social networks do not have normal social relationships, and the relational network structure formed around spammers is relatively special. Therefore, from the perspective of relational networks, spammers in social networks can be well detected.
Krestel et al.~\cite{krestel2008using} proposed an algorithm to identify spammers from the collaborating systems by employing a spam score propagating technique. This method takes advantage of the characteristics that spammers' suspicion will spread in social networks, and then uses the spread on the graph model to find spammers in social networks.
Bhat et al.~\cite{bhat2013community} found that similar to normal users, spammers in social networks can also form the spammer community. Therefore, they extracted user interaction graphs from user history behaviour and found overlapping community graphs among them. After manually marking a part of spammer nodes, they calculate the community relationship between each node to be identified and the marked node to classify unknown nodes.
Brophy et al.~\cite{brophy2017collective} tried to construct a topological structure graph for each relation on the social network, using complex network features such as \emph{Triangle Count}~\cite{schank2007algorithmic}, K-Core~\cite{alvarez2006large}, PageRank~\cite{page1999pagerank}, connected components~\cite{pemmaraju2003computational} and other topological features to construct the features of spammers on social networks. They assume that spammers occupy a very important position in each network topology graph.

%The network topological structure, which is the implicit attribute of social network has raised researchers attention.
%In order to exploit the network topological structure, graph-based methods have been proposed in recent literature~\cite{fakhraei2015collective,Fayazi15SIGIR,fakhraei2016adaptive,jeong2016follow}.
%Specifically,
%graph-based features are extracted by converting relations into a directed graph $\mathcal{G}$, where the vertices $\mathcal{V}$ represent the users and the edges $\mathcal{E}$ represent interactions among users.
%When there exist multiple types of relations, a graph is usually generated for each of them: $\{\mathcal{G}_1, \ldots, \mathcal{G}_m\}$ for $m$ kinds of relations. Then, each graph uses graph analytic methods to extract graph-based features.
%This converts a directed graph into either a numerical or categorical feature matrix for each kind of relation.
%Existing literature has defined many graph analytic methods~\cite{tu2017transnet}, 
%i.e., \emph{Triangle Count}~\cite{schank2007algorithmic},
%\emph{k-core}~\cite{alvarez2006large}, 
%\emph{Graph Coloring}~\cite{jensen2011graph},
%\emph{Page Rank}~\cite{page1999pagerank},
%and \emph{Weakly Connected Components}~\cite{pemmaraju2003computational}.

Nevertheless, graph-based methods are effective under the assumption that the data is homogeneous, i.e., different types of relations are required to be modeled separately. Unfortunately, this assumption ignores the interactions among different types of relations.
Sequence-based methods alleviate the limitation of graph analytic methods to a certain extent, as it models all relations together.
In more detail,
sequence-based features are extracted by converting
different types of relations into a user-wise sequence,
and the length of each sequence depends on the user.
The sequence of each user is then fed into a feature extraction function
to convert the sequence of user into a feature vector.
For example, Wang et al.~\cite{wang2017clickstream} proposed a clickstream models to calculate the distance between each clickstream traces(i.e., sequences of click events from users). They assume that spammers and normal users exhibit different click transition patterns and focus their energy on different activities.
Fakhraei et al. proposed \emph{Sequential k-gram Features}~\cite{fakhraei2015collective} which considers the activity order of users by counting the frequency of each length $k$ sub-sequences for each user.
However, Fakhraei et al. only considered the situation when $k=2$, 
for the reason that the large $k$ will cost huge computing spaces.
Subsequently, Peng et al. announced that
\emph{Mixture of Markov Models}~\cite{peng2004augmenting} can be used to overcome the limitation of small $k$ in $k$-gram models by identifying a small set of important sequence from a long sequence chains.
Nevertheless, 
Mixture of Markov Models only considered the short-term information within users' relational sequences.
Overall,
the sequence-based features extracted in the literature 
can not take long-term interactions along with the short-term information into consideration at the same time.

\section{Problem Formulation}
\label{sec:preliminary}
In this section, we begin by illustrating individual-level and union-level dependency of users' interaction sequences, which actually motivates our work. Then, we formulate the social spammer detection problem and introduce the overall framework of our Multi-level Dependency Model.

\subsection{Motivation}
\label{sec:motivation}
\begin{figure*}[htbp]
	\centering
	\subfigure[The relational sequence of normal user A]{
		\includegraphics[width=0.9\linewidth]{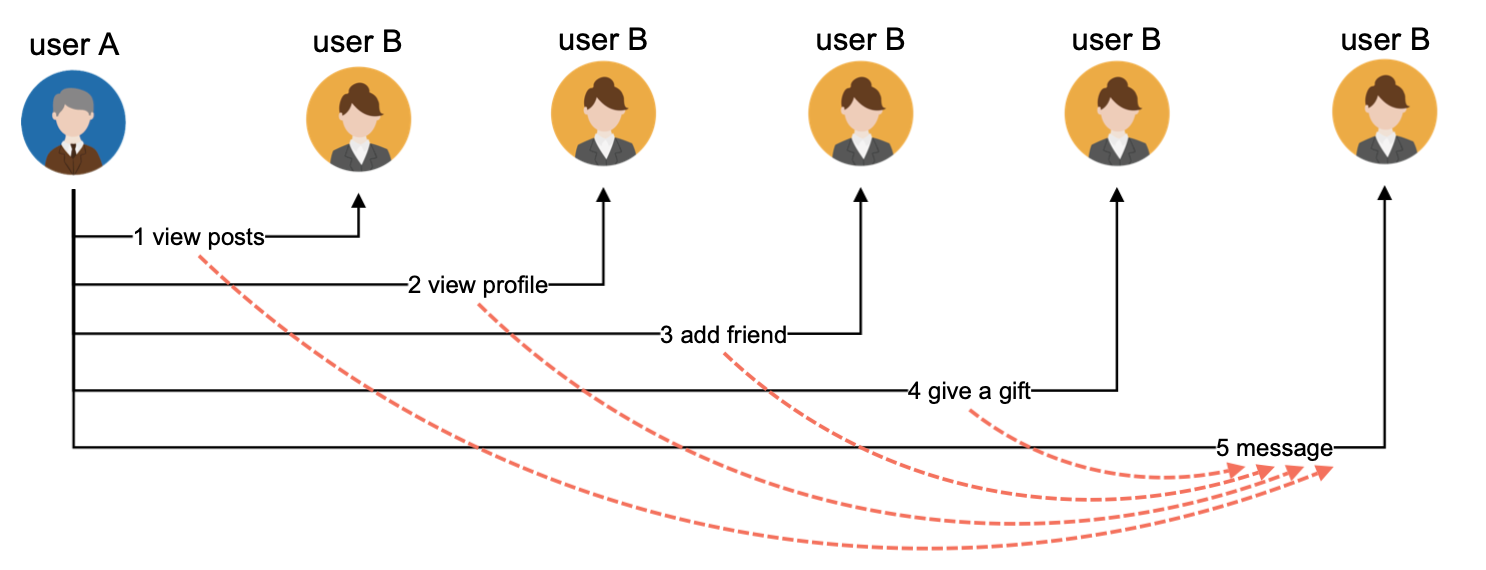}
		\label{fig:normal_indi}}
	\subfigure[The relational sequence of spammer E]{
		\includegraphics[width=0.93\linewidth]{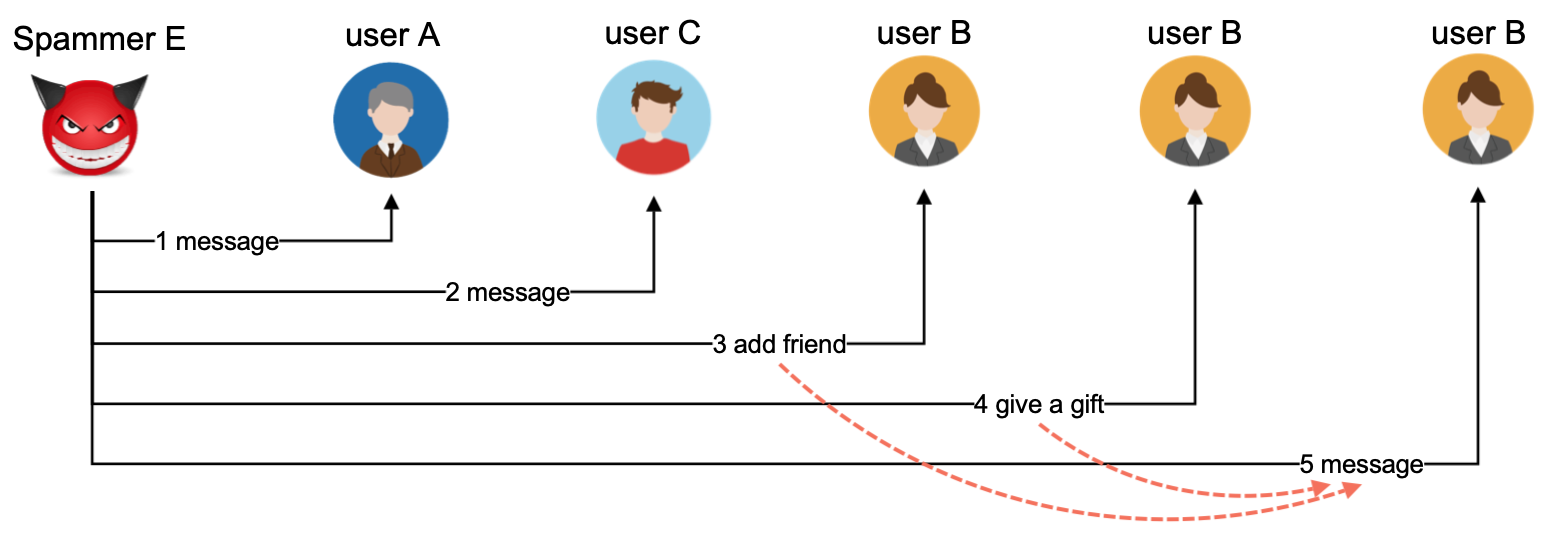}
		\label{fig:spam_indi}}
\caption{Examples of individual-level dependency among the relational sequence. Normal users may be involved in one of the given sequence of relations in (a), e.g., \emph{view posts} followed by \emph{send message}. Spammer can only imitate one or two relation behaviours from normal users (e.g.,\emph{add friend} followed by \emph{send message}). But it is impossible for spammers to completely imitate the all behaviour sequences in (a), because spammers always have their own malicious purposes.}
\label{fig:toyexample1}
\end{figure*}

\begin{figure*}[htbp]
	\centering
	\subfigure[Relational sequence of normal user A]{
		\includegraphics[width=0.9\linewidth]{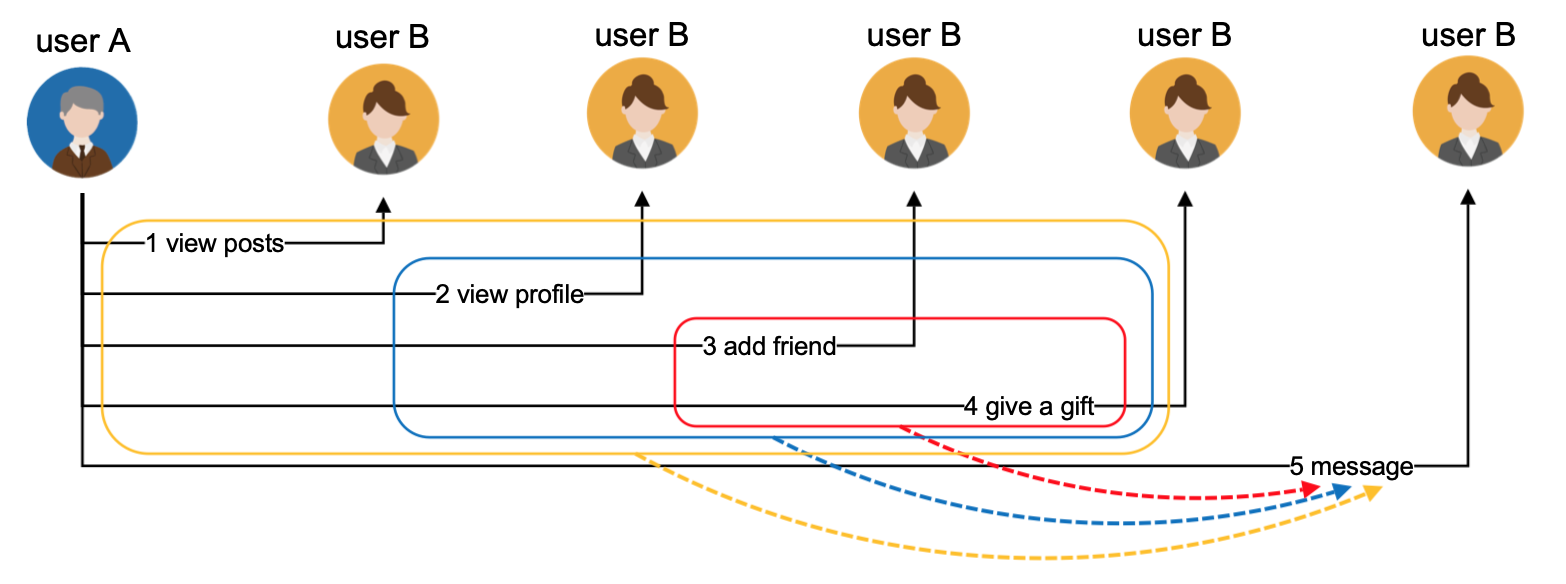}
		\label{fig:normal_union}}
	\subfigure[Relational sequence of spammer E]{
		\includegraphics[width=0.9\linewidth]{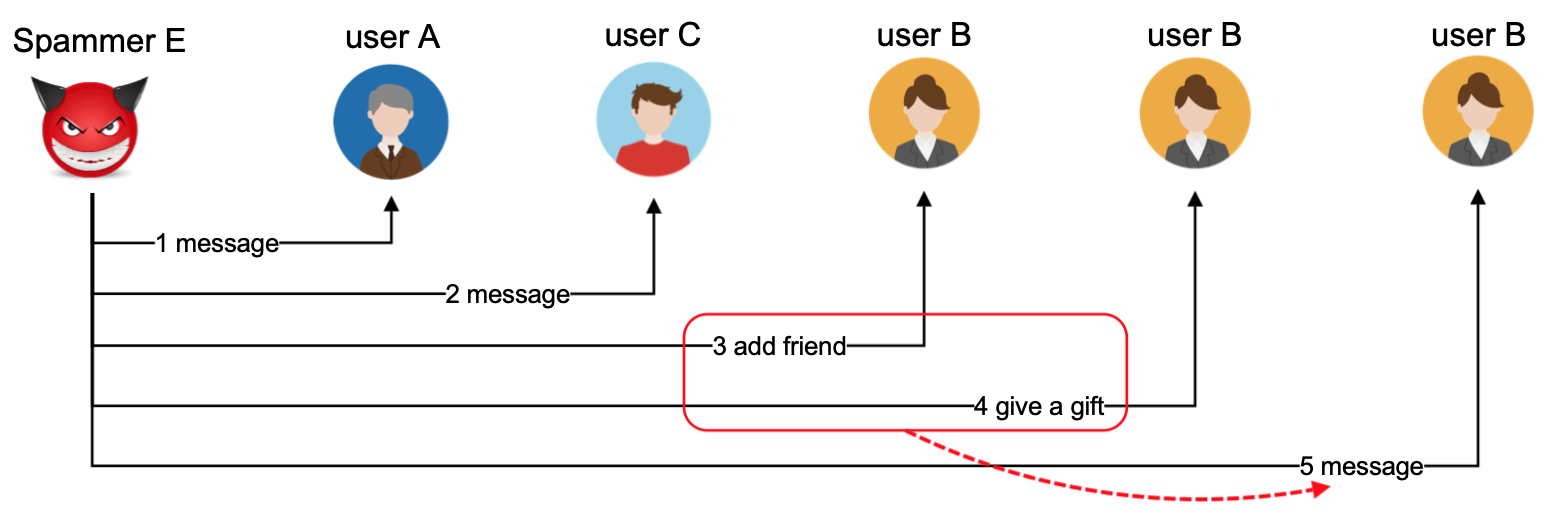}
		\label{fig:spam_union}}
\caption{Examples of union-level dependency among the relational sequence. The union-level dependency can somehow capture the collective influence among a union of relations that the spammer performs. For example, the spammer is more likely to add friend, giving a gift and sending a message together than adding friend, giving a gift or sending a message individually.}
\label{fig:toyexample2}
\end{figure*}

Inspired by the sequential recommendation method in e-commence~\cite{yu2019multi,zhou2018personalized,tang2018personalized}, the most recent $n$ items that a user bought play an important role in the prediction of the next item user wants to buy.
In the context of multi-relational social network, 
we assume that the ultimate purpose of spammer is \emph{sending messages} to as much users as possible, aiming to spread false information.
Hence, the difference between spammer and normal user in the most recent $n$ relations before \emph{sending a messages}, instead of whole day's relational sequence (long-term) can be detected as the short-term dependency of their relational sequences.
Moreover, we exploit the users' short-term relational sequence in terms of individual-level and union-level dependency. An example is shown in Fig.~\ref{fig:toyexample1} and Fig.~\ref{fig:toyexample2} for illustration.

Fig.~\ref{fig:normal_indi} shows a relational sequence of a normal user A.
Firstly, user A views user B's posts, and then view the profile of user B. 
User A will add user B as friend if he/she is interested in user B. 
After that, they start to have further interaction with sending messages or even give a gift. From the individual point of view, every relation user A performed may result in the \emph{message} happen as the dotted lines with arrows indicating in Fig.~\ref{fig:normal_indi}. 
However, it is easy for spammer to imitate.
As shown in Fig.~\ref{fig:spam_indi}, spammer E performed \emph{add friend} and \emph{give a gift} before \emph{message} to hide its behaviours from being detected.

Although spammer is capable of imitating the normal user's individual-level pattern, spammer may fail to imitate the union-level pattern of the relational sequence.
We can illustrate this by another example in Fig.~\ref{fig:normal_union}.
User A does a \emph{message} relation to user B preceded by different combination of the previous relations.
Spammer E in Fig.~\ref{fig:spam_union} may somehow copy the simplest union-level, however, it is difficult for it to imitate a complicated union-level. Thus, in order to improve the performance of social spammer detection, we shall exploit the short-term dependency of user's relational sequence from both individual-level and union-level.

With the modeling of short-term dependency, we also exploit the long-term dependency among users' relational sequences.
This is mainly because only considering short-term (only a few relations user performed) is biased, as long-term (user's whole day or even whole week's relational sequence) may expose user's general behaviour and intention.
Consequently, our work exploits users' relational sequences in terms of long-term dependency of their relational sequences along with short-term dependency from both individual-level and union-level.

\subsection{Framework Overview}
\label{sec:formulation}

Let $\mathcal{U}$ be the set of $N$ users, $u\in \mathcal{U}$ be a user. Note that we will also use $u_i, u_j \in \mathcal{U}$ to denote different users. Suppose there are $M$ types of relations among users, denoted as $\mathcal{R}=\{r_1,\cdots, r_M\}$. Specifically, $M=7$ in our paper indicates seven relations including ``add friend'', ``message'', ``give a gift'', ``view profile'', ``pet game'', ``meet-me game'', ``report abuse''. We represent each user as a relational sequence $u=\langle s^u_1, \cdots, s^u_t, \cdots, s^u_T \rangle$, where $s^u_t\in \mathcal{R},~1\leq t \leq T$ and the index $t$ denotes the order in which one type of relation is used by $u$. The target of spammer detection is to estimate the likelihood that every user belongs to the spammer class, denoted as $P(y_u=\text{spammer}|u)$, where $y_u$ is the label of $u$ within the domain $\{\text{normal user}, \text{spammer}\}$. For simplicity, we let $\phi_u=P(y_u=\text{spammer}|u)$ and it is defined as:
\begin{equation}
\phi_u=\mathbf{F}(u, n)\cdot \sum_{r_{m} \in u}{\boldsymbol{m}_{r_m}^{\top}},
\label{eq:predictive model}
\end{equation}
where $\boldsymbol{m}_{r_m}\in\mathbb{R}^d$ is the embedding of relation, $r_m\in\mathcal{R}$,
$n$ is the selected most recent $n$ relations for short-term modeling.
$\mathbf{F}(u, n)$ is the output of proposed \emph{MDM}.

\begin{figure}[htbp]
	\centering
	\includegraphics[width=0.8\linewidth]{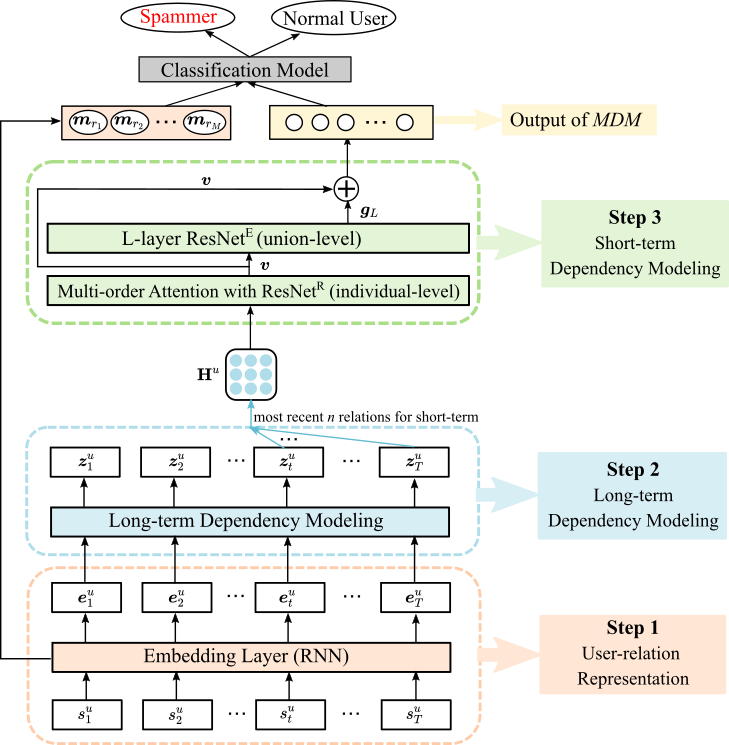}
	\caption{Framework of Multi-level Dependency Model (\emph{MDM}).}
	\label{fig:framework}
\end{figure}

The overall architecture of proposed \emph{MDM} consists of 
\emph{User-relation Representation}, 
\emph{Long-term Dependency Modeling}
and \emph{Short-term Dependency Modeling} as shown in Fig.~\ref{fig:framework}. \emph{MDM} first uses skip-gram with Recurrent Neural Network for representing user-relation into vector embedding. As shown in the bottom layer of Fig.~\ref{fig:framework}, the input of this layer is one user's relational sequence $u=\langle s^u_1, \cdots, s^u_t, \cdots, s^u_T \rangle$. While the output is the $d$-dimensional latent vector of the input relational sequence.

After that, \emph{MDM} models long-term order constraint over the whole user-relation vector embeddings with a \emph{Long-term Dependency Modeling} layer. The \emph{Long-term Dependency Modeling} layer maps the whole user-relation vectors into a sequence of hidden vectors. With the output of \emph{User-relation Representation} layer, \emph{Long-term Dependency Modeling} generates the most recent $n$ relations latent vectors as matrix $\mathbf{H}^u$.
More importantly, we design one further step to input $\mathbf{H}^u$ to an attention layer, from which the short-term dependency is learned by \emph{Short-term Dependency Modeling}, as shown in Fig.~\ref{fig:framework}. 
This is similar to the most recent $n$ items containing the potential intentions and preferences of the user, which can predict users' next behaviour. Finally, both long-term and short-term hidden information are extracted as embedding features and fed into a classification model for the spammer detection task.
All the notations are listed in Table~\ref{tab:notation}.

\begin{table}[H]
	\centering \scriptsize
	\caption{Summary of notations}
	\label{tab:notation}
	\scalebox{0.92}{	
	\begin{tabular}{l|l}
		\toprule
		Notation  &  Description \\
		\midrule
		$\mathcal{L},\mathcal{S},\mathcal{U}$ &  Set of normal users, spammers and all users respectively, $\mathcal{L}\cup \mathcal{S}=\mathcal{U}$\\
		$u$ &  User's relational sequence, $u=\langle s^u_1, \cdots, s^u_t, \cdots, s^u_T \rangle$, $u\in \mathcal{U}$\\   
		$\mathcal{R}$ &  Set of relations, $\mathcal{R}=\{r_1,\cdots,r_m,\cdots r_M\}$\\  
		$\mathbf{F}(u,n)$ &  Output of \emph{MDM}\\ 
		$\boldsymbol{m}_{r_m}$ &  The embedding of relation, $\boldsymbol{m}_{r_m}\in\mathbb{R}^d$\\
		$\boldsymbol{e}_t^u$ &  The user-relation representation for position $t$ in $u$, $\boldsymbol{e}_t^u\in\mathbb{R}^d$\\
		$\boldsymbol{z}_t^u$ &  Output of \emph{Long-term Dependency Modeling}, $\boldsymbol{z}_t^u\in\mathbb{R}^d$\\
		$u_n$ &  Set of most recent happened $n$ relations, $u_n=\langle s_{T-1}^u, s_{T-2}^u,\cdots, s_{T-n}^u\rangle$\\
		$\mathbf{H}^u$ &  The most recent $n$ relations' outputs from \emph{Long-term Dependency Modeling} layer\\
		$k,~L$ &  Numbers of layers for $ResNet^R$ and $ResNet^E$ respectively\\
		$\mathbf{H}_1^u, \mathbf{H}_2^u,\cdots,\mathbf{H}_k^u$ &  Hidden status of $ResNet^R$\\
		$\boldsymbol{v}_l$ &  High-order features for each layer of $ResNet^R$, $\boldsymbol{v}_l\in \mathbb{R}^d~(0\leq l\leq k)$\\
		$\boldsymbol{h}_{i:}^{k}$ &  Corresponding $i$-th row of matrix $\mathbf{H}_k^u$\\
		$\alpha_i^k$ &  Weight scale for $\boldsymbol{v}_k$\\
		$[\boldsymbol{v}_0, \boldsymbol{v}_1,\cdots, \boldsymbol{v}_k]^{\top}$ &  Set of aggregated high-order features\\
		$\boldsymbol{v}$ &  Output of \emph{Multi-order Attention with $ResNet^R$ (indiviual-level)} layer\\
		$\beta$ & Attention weight vector,  $\beta \in \mathbb{R}^{k+1}$\\
		$\boldsymbol{g}_L$ &  Output of \emph{L-layer $ResNet^E$ (union-level)} layer\\
		$\Theta$ &  Parameters for optimizing, including:\\&  $W_{LSTM}$ for long-term modeling; $\mathbf{W}_k, \boldsymbol{b}_k$ for $ResNet^R$;\\& $\mathbf{\omega}_1, \mathbf{\omega}_2, c_1, c_2, \mathbf{\varphi}_1, \mathbf{\varphi}_2, b_1, b_2$ for attention model and $\mathbf{W}_L, \boldsymbol{b}_L$ for $ResNet^E$.\\
		\bottomrule
	\end{tabular}
	}
\end{table}

\section{Multi-level Dependency Model}
\label{sec:method}
This section discusses the three components of \emph{Multi-level Dependency Model} (\emph{MDM}) in more details.

\subsection{User-relation Representation}
\label{sec:representation}
Before exploiting the hidden information behind user's relational sequence,
our prior problem is to model it.
In order to reveal relation's sequential characteristics which is implied in the relational sequence,
it is necessary to find an effective representation method to directly learn high-quality user-relation vectors from the users' relational sequences. 
We apply the skip-gram with Recurrent Neural Network~\cite{lipton2015critical} 
to generate user-relation representations by exploiting the users' relational sequences.

Specifically, 
given a relation $r_m~(1\leq m\leq M)$ and a user's relational sequence $u=\langle s^u_1, \cdots, s^u_t, \cdots, s^u_T \rangle$, we denote the likelihood of $s^u_t = r_m$ as
\begin{equation}
    P(r_m|~t, u)=\frac{\exp(\varepsilon(r_m,s^u_t))}{\sum^{M}_{m'=1}{\exp(\varepsilon(r_{m'},s^u_t))}},
\end{equation}
where $\varepsilon(r_m,s^u_t)=\boldsymbol{m}_{r_m} \cdot \boldsymbol{e}_{t}^{{u}^{\top}}$, $\boldsymbol{m}_{r_m}\in \mathbb{R}^d~(1\leq m\leq M)$ is the latent vector for each relation in $\mathcal{R}$, 
and $\boldsymbol{e}_{t}^{u}\in \mathbb{R}^d$ is the user-relation representation for position $t~(1\leq t\leq T)$ in $u$. 
To obtain the embedding $\boldsymbol{m}_{r_m}$ and $\boldsymbol{e}_{t}^{u}$, the \emph{Embedding layer} implemented by RNN optimize the objective function as follows:
\begin{equation}
	    \max_{\boldsymbol{m}_{r_m},\boldsymbol{e}_{t}^{u}} \sum_{m=1}^{M}\sum_{t=1}^{T}\log P(r_m|~t, u)
		=\max_{\boldsymbol{m}_{r_m},\boldsymbol{e}_{t}^{u}} \sum_{m=1}^{M}\sum_{t=1}^{T}\log\frac{\exp[\boldsymbol{m}_{r_m} \cdot \boldsymbol{e}_{t}^{{u}^{\top}}]}{\sum^{M}_{m'=1}{\exp[\boldsymbol{m}_{r_{m'}} \cdot \boldsymbol{e}_{t}^{{u}^{\top}}]}},
	\label{eq:representation}
\end{equation}
where $T$ is the length of relational sequence $u$, and $M$ is the number of relations. 

\subsection{Long-term Dependency Modeling}
\label{sec:lstm}
To model the long-term dependency of user's relational sequence on multi-relational social networks,
we apply a standard LSTM~\cite{hochreiter1997long} 
as in Fig.~\ref{fig:framework}
over the whole relational sequence.
For each $u\in \mathcal{U}$ we can get a user-relation representation from Eq.~\eqref{eq:representation}, denoted as
$\{\boldsymbol{e}^{u}_1,\cdots,\boldsymbol{e}^{u}_t,\cdots,\boldsymbol{e}^{u}_{T}\}$
, where $\boldsymbol{e}^{u}_t$ denotes the $d$-dimensional latent vector of position $t$.
Given the user-relation representation for user $u$ from the last \emph{User-relation Representation} layer, 
we can obtain a sequence of hidden vectors $\{\boldsymbol{z}^{u}_1,\cdots,\boldsymbol{z}^{u}_t,\cdots,\boldsymbol{z}^{u}_{T}\}$
by recurrently inputting $\boldsymbol{e}^{u}_t~(1\leq t\leq T)$ into LSTM, i.e.,
\begin{equation}
	\boldsymbol{z}^{u}_t=\text{LSTM}(\boldsymbol{e}^{u}_t,\boldsymbol{z}^{u}_{t-1},W_{LSTM}),
\label{eq:long-term}
\end{equation}
where $\text{LSTM}$ is the output function of Long Short-Term Memory, $W_{LSTM}$ contains the weight parameters
and we set $\boldsymbol{z}^{u}_0 = \boldsymbol{0}$.

Through this stage, 
the \emph{Long-term Dependency Modeling} in Fig.~\ref{fig:framework}
outputs a sequence 
$\{\boldsymbol{z}^{u}_1,\cdots,\boldsymbol{z}^{u}_t,\cdots,\boldsymbol{z}^{u}_{T}\}$
for the next multi-order attentive relation modeling stage.
Since only capturing long-term dependency is not sufficient, 
as it neglects the importance of adjacent relation within the sequence.
In next section, 
we will illustrate how to augment long-term dependency with short-term dependency in terms of individual-level and union-level.

\subsection{Short-term Dependency Modeling}
\label{sec:short-term}
In this section, we will discuss how to extend general user's embedding with short-term dependency over a small set of the most recent $n$ happened relations,
which can be denoted as $u_n=\langle s_{T-1}^u, s_{T-2}^u,\cdots, s_{T-n}^u\rangle$.

As can be seen from Fig.~\ref{fig:framework} the \emph{MDM} applies $ResNet$ to learn high-order non-linear interactions among the short-term dependency of $u_n$.
\emph{MDM} instantiates two residual networks with a fully connected multi-layer perceptron,
i.e., $k$-layer $ResNet^R$ for individual-level and $L$-layer $ResNet^E$ for union-level, respectively.

\subsubsection{Individual-level}
\label{sec:individual}
\begin{figure*}[htbp]		
    \centering
	\includegraphics[width=0.8\linewidth]{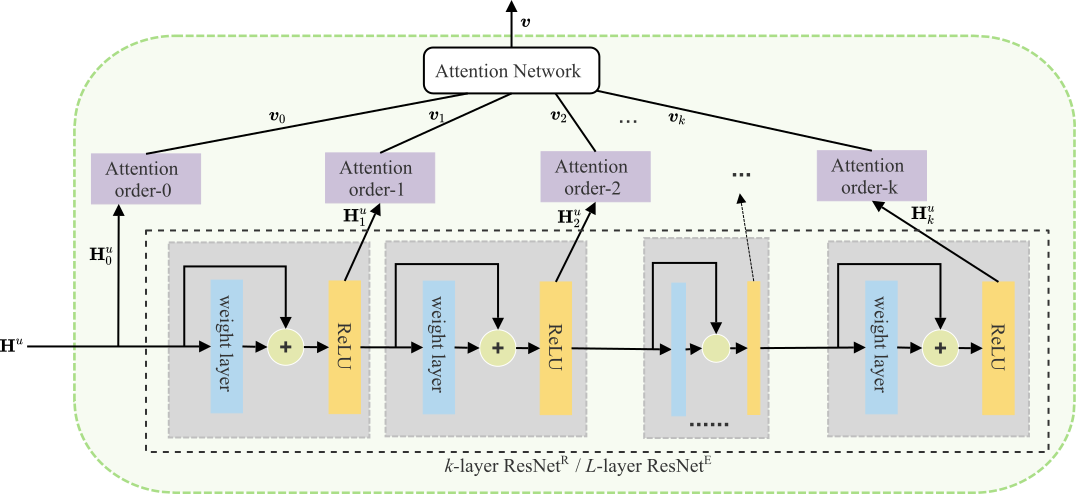}
    \caption{Multi-order Attention Network.}
    \label{fig:attention}
\end{figure*}

As shown in Fig.~\ref{fig:toyexample1}, the individual-level dependency among the relational sequence may result in different subsequent relations for the normal user and spammer, respectively.
In other words, exploiting the individual-level dependency will definitely benefit the spammer detection.
Thus, we take the most recent $n$ relations' outputs from the last \emph{Long-term Dependency Modeling} layer, 
denoted as $\mathbf{H}^u \in \mathbb{R}^{n\times d}$:
\begin{equation}
\label{eq:h}
    \mathbf{H}^u=\begin{bmatrix}
    \boldsymbol{z}_{T-1}^u\\ 
    \boldsymbol{z}_{T-2}^u\\ 
    \vdots\\ 
    \boldsymbol{z}_{T-n}^u
    \end{bmatrix},
\end{equation}
where $\boldsymbol{z}_{T-1}^u$ is produced by Eq.~\eqref{eq:long-term} with $t=T-1$.
And we use $\mathbf{H}^u$ as the input of \emph{Multi-order Attention with $ResNet^R$ (individual-level)} layer shown in Fig.~\ref{fig:framework}. 
Then, we propose an attention mechanism to aggregate high-order features of individual level dependency as show in Fig.~\ref{fig:attention}.

With the input embedding $\mathbf{H}^u$, 
the \emph{Multi-order Attention with $ResNet^R$ (individual-level)} layer is instantiated with a $k$-layer residual network $ResNet^R$,
i.e., $ResNet(k, \mathbf{H}^u)$.
Then, we can obtain the output of a sequence of hidden status as
\begin{equation}
	\begin{split}
		\mathbf{H}_1^u&=\text{ReLU}(\mathbf{H}^u\mathbf{W}_1+\mathbf{b}_1+\mathbf{H}^u)\\
		\mathbf{H}_2^u&=\text{ReLU}(\mathbf{H}_1^u\mathbf{W}_2+\mathbf{b}_2+\mathbf{H}_1^u)\\
		&\cdots\cdots\\
		\mathbf{H}_k^u&=\text{ReLU}(\mathbf{H}_{k-1}^u\mathbf{W}_k+\mathbf{b}_k+\mathbf{H}_{k-1}^u),
	\end{split}
\label{eq:resnet}
\end{equation}
where $\text{ReLU}$ is the activation function for \textit{rectifier linear unit}, $\mathbf{H}_k^u\in \mathbb{R}^{n\times d}$ is the high-order features generated at $k$-th layer of $ResNet^R$, $k$ denotes the maximum number of residual layers.
$\mathbf{W}_k \in \mathbb{R}^{d\times d}$ and $\mathbf{b}_k \in \mathbb{R}^d$ denote weight matrix and bias vector, respectively.

The sequence of hidden status from $ResNet^R$, i.e., $\{\mathbf{H}_1^u, \mathbf{H}_2^u,\cdots,\mathbf{H}_k^u\}$, can capture potential high-order interactions between partial fields in relation embedding,
which helps us to discriminate the significance of relations in different levels with respect to a given relational sequence.
Besides the high-order features, we also keep the raw embedding $\mathbf{H}^u$, resulting a set of extended encoded features $\{\mathbf{H}_0^u, \mathbf{H}_1^u,\cdots,\mathbf{H}_k^u\}$, where $\mathbf{H}_0^u=\mathbf{H}^u$.

To aggregate $\{\mathbf{H}_0^u, \mathbf{H}_1^u,\cdots,\mathbf{H}_k^u\}$, we use a soft attention model.
We denote $\boldsymbol{v}_l \in \mathbb{R}^{d}~(0\leq l\leq k)$ as the contextual embedding for each layer, which can be generated by the soft attention model~\cite{yu2019multi} as follows.
\begin{equation}
	\begin{split}
		\boldsymbol{v}_0&=\sum_{i=1}^{n}\alpha_{i}^{0}\cdot \boldsymbol{h}_{i:}^{0}\\
		\boldsymbol{v}_1&=\sum_{i=1}^{n}\alpha_{i}^{1}\cdot \boldsymbol{h}_{i:}^{1}\\
		&\cdots\cdots\\
		\boldsymbol{v}_k&=\sum_{i=1}^{n}\alpha_{i}^{k}\cdot \boldsymbol{h}_{i:}^{k},
	\end{split}
\label{eq:contextual embedding}
\end{equation}
where $\boldsymbol{h}_{i:}^{k}~(1\leq i\leq n)$ is the corresponding $i$-th row of matrix $\mathbf{H}_k^u$.
And weight scale $\alpha_{i}^{k}$ is normalized by a softmax layer on the attention scores, $\sum_{i=1}^{n}\alpha_{i}^{k}=1$.
We utilize a network with two-layers to calculate the attention scores with Eq.~\eqref{eq:att}.
\begin{equation}
\begin{split}
	\alpha_{i}^k&=\mathbf{\omega}_1\text{tanh}(\mathbf{\omega}_2\boldsymbol{h}_{i:}^{k}+c_1)+c_2\\
	\alpha_{i}^k&=\frac{\text{exp}(\alpha_{i}^k)}{\sum_{i'=1}^{n}\text{exp}(\alpha_{i'}^k)},
\end{split}
\label{eq:att}
\end{equation}
where $\mathbf{\omega}_1, \mathbf{\omega}_2$ are the shared weight matrices for attention layer.
Then, the final contextual embedding
of short-term dependency with the most recent $n$ relations is
\begin{equation}
\begin{split}
	&\boldsymbol{v}=\beta[\boldsymbol{v}_0, \boldsymbol{v}_1,\cdots, \boldsymbol{v}_k]^{\top}\\
	\beta=\text{softmax}&(\mathbf{\varphi}_1\text{tanh}(\mathbf{\varphi}_2[\boldsymbol{v}_0, \boldsymbol{v}_1,\cdots, \boldsymbol{v}_k]^{\top}+b_1)+b_2 ),
\end{split}
\label{eq:softatt}
\end{equation}
where $\mathbf{\varphi}_1, \mathbf{\varphi}_2$ are the weight matrices for attention layer 
and $\beta \in \mathbb{R}^{k+1}$ is the attention weight vector.
According to Eq.~\eqref{eq:contextual embedding} and Eq.~\eqref{eq:softatt},
we can obtain the final contextual embedding $\boldsymbol{v}$ on individual-level as Eq.~\eqref{eq:individual}
\begin{equation}
	\boldsymbol{v}=\beta\cdot\sum_{l=0}^{k}\sum_{i=1}^{n} \alpha_{i}^{l}\cdot \boldsymbol{h}_{i:}^{l}.
\label{eq:individual}
\end{equation}

\subsubsection{Union-level}
\label{sec:union}
In addition to the individual-level dependency,
we also exploit the union-level of short-term dependency among the relational sequences of users.
As shown in Fig.~\ref{fig:toyexample2}, 
although spammers might imitate some individual-level patterns from the normal users' relational sequences,
it is difficult for them to imitate a complicated combination of normal relations, i.e., union-level dependency.
Therefore, we argue that individual-level and union-level dependency can be complementary to tackle users' short-term relational sequences.

The union-level dependency can be conceptually understood by estimating the probability of an associate rule $X\to Y$, where $X$ is the most recent $n$ relations of one user and $Y$ is the subsequent relation to be performed.
In particular, we combine attention network and residual network to represent relation set $X$.
Specifically, 
we use the embedding features from individual-level as the input to a multi-layer perceptron with residual structure, instantiated as $ResNet^E$.
With the input $\boldsymbol{v}$ given by Eq.~\eqref{eq:individual}, $ResNet^E$ outputs the representation of union-level dependencies as follows. 
\begin{equation}
    \begin{split}
		\boldsymbol{g}_1&=\text{ReLU}(\boldsymbol{v}\mathbf{W}_1+\mathbf{b}_1+\boldsymbol{v})\\
		\boldsymbol{g}_2&=\text{ReLU}(\boldsymbol{g}_1\mathbf{W}_2+\mathbf{b}_2+\boldsymbol{g}_1)\\
		&\cdots\cdots\\
		\boldsymbol{g}_L&=\text{ReLU}(\boldsymbol{g}_{L-1}\mathbf{W}_L+\mathbf{b}_L+\boldsymbol{g}_{L-1}),
	\end{split}
\label{eq:gL}
\end{equation}
where $\mathbf{W}_L \in \mathbb{R}^{d\times d}$ and $\mathbf{b}_L \in \mathbb{R}^d$ denote weight matrix and bias vector, respectively.

\subsection{Objective Function}
\label{sec:op}
To concatenate individual-level features with union-level features, 
we formulate $\mathbf{F}(u, n)$ in Eq.~\eqref{eq:predictive model} for user with relational sequence $u\in\mathcal{U}$ as:
\begin{equation}
    \mathbf{F}(u, n) = \boldsymbol{v}+\boldsymbol{g}_L,
\label{eq:concatenation}
\end{equation}
where $\boldsymbol{v}$ is given by Eq.~\eqref{eq:individual} and $\boldsymbol{g}_L$ is given by Eq.~\eqref{eq:gL}.
$\mathbf{F}(u, n)$ is then the embedding of context information integrated at both individual-level and union-level.
Afterwards, the predictive model Eq.~\eqref{eq:predictive model} can be extended as:
\begin{equation}
	\phi_u=\mathbf{F}(u, n)\cdot \sum_{r_{m} \in u}{\boldsymbol{m}_{r_m}^{\top}}\\
	=(\boldsymbol{v} + \boldsymbol{g}_L)\cdot \sum_{r_{m} \in u}\boldsymbol{m}_{r_m}^{\top}.
\label{eq:final predic}
\end{equation}
The spammers will have relative larger values of Eq.~\eqref{eq:final predic} than normal users, i.e., $\phi_{u_i}>\phi_{u_j}$. 

The predictive model $\phi_u$ can be fitted by optimizing the underlying parameters $\Theta$ that is from $W_{LSTM}$ in Eq.~\eqref{eq:long-term}, $ResNet^R$ in Eq.~\eqref{eq:resnet}, $ResNet^E$ in Eq.~\eqref{eq:gL} and soft attention model in Eq.~\eqref{eq:individual}.
Let $\mathcal{S}$ represents the set of spammers' relational sequences and $\mathcal{L}$ denotes the set of normal users' relational sequences, i.e., $\mathcal{U}=\mathcal{S}\cup \mathcal{L}$.
With the inputs of user-relation sequences $u=\langle s^u_1, \cdots, s^u_t, \cdots, s^u_T \rangle$, $\Theta$ can be obtained by optimizing the following objective function:
\begin{equation}
	\mathop{\arg\min}_{\Theta}\sum_{u_i \in \mathcal{S}}\sum_{u_j \in \mathcal{L}}-I(\phi_{u_i}, \phi_{u_j}) + \frac{\lambda}{2}||\Theta||_F^2,
\label{eq:loss function}
\end{equation}
where $I(\cdot,\cdot)$ is an indicator function that equals 1 for $\phi_{u_i}>\phi_{u_j}$, otherwise equals 0,
$||\cdot||_F^2$ represents \emph{Frobenius} norm weighted with a hyper-parameter $\lambda$.
We use Adam optimizer~\cite{kingma2014adam} to optimize the objective function~\eqref{eq:loss function} and produce the optimal $\Theta$. 
The pseudocode of leveraging \emph{MDM} for social spammer detection is presented in Alg.~\ref{alg:mdm}.
\begin{algorithm}[ht]
\caption{Algorithm of Leveraging \emph{MDM} for Social Spammer Detection} 
\label{alg:mdm}
		\begin{algorithmic}[1]
		\Require Labeled set $\mathcal{U}=\mathcal{S}\cup \mathcal{L}$ includes all users' relational sequences, and each user's relational sequence is $u=\langle s^u_1, \cdots, s^u_t, \cdots, s^u_T \rangle$; Number of relations for short-term $n$; Embedding size $d$; Number of layers $k$.
		\Ensure Users' label: \{Spammer, Normal user\}.
		\Procedure{MDM}{$\mathcal{U},n,d,k$}\\
		\quad~\algorithmicrepeat 
		\For {each $u\in \mathcal{U}$}
        	\State Compute $\boldsymbol{e}^u_t$ and $\boldsymbol{m}_{r_m}$ via Eq.~\eqref{eq:representation}; \algorithmiccomment{\textit{Step 1}}
    		\State Compute the long term embedding $\boldsymbol{z}^u_1,\cdots,\boldsymbol{z}^u_T$ by Eq.~\eqref{eq:long-term}; \algorithmiccomment{\textit{Step 2}}
    		\State Compute $\mathbf{H}^u$ via Eq.~\eqref{eq:h}; \Comment{\textit{Step 3, individual-level}}
    		\State Compute $\{\mathbf{H}_0^u,\mathbf{H}_1^u,\cdots,\mathbf{H}_k^u\}$ via Eq.~\eqref{eq:resnet}; \algorithmiccomment{$\mathbf{H}_0^u=\mathbf{H}^u$}
    		\State Compute $[\boldsymbol{v}_0, \boldsymbol{v}_2,\cdots, \boldsymbol{v}_k]^{\top}$ by $\{\mathbf{H}_0^u, \mathbf{H}_1^u,\cdots,\mathbf{H}_k^u\}$  via Eq.~\eqref{eq:contextual embedding};
    		\State Compute the embedding of individual level $\boldsymbol{v}$ via Eq.~\eqref{eq:softatt};
    		\State Compute the embedding of union level $\boldsymbol{g}_L$ by Eq.~\eqref{eq:gL}; \algorithmiccomment{\textit{union-level}}
    		\State Compute $\mathbf{F}(u, n)$ via Eq.~\eqref{eq:concatenation}; \algorithmiccomment{\textit{Concatenate}}
    		\State Update the parameter set $\Theta$ in~\eqref{eq:loss function} by Adam algorithm;
    	\EndFor
    	\State \textbf{until} converge.
    	\EndProcedure
    	\Procedure{Prediction}{$MDM(\cdot)$, $\mathcal{U}$}
	        \State Take the output of \emph{MDM}, $\mathbf{F}(u, n)$, as the feature for each user $u$;
	        \State Use classification model to classifier spammers and normal users;
    	\EndProcedure
\end{algorithmic}
\end{algorithm}
\footnote{The details of step 1, 2 and 3 can be found in Fig.~\ref{fig:framework}}

\subsection{An Illustrating Example}
\label{sec:example}
In order to better understand the overall process of our proposed \emph{MDM}, we give a simple example in this section.

The input of our method is the user's relational sequence collected from Tagged.com, e.g., $u=\langle 5,5,5,4,4,3,5,4,4 \rangle$. When inputting $u$ to our \emph{MDM} method, \emph{User-relation Representation} layer outputs two components: One is a $d$-dimensional latent vector $\boldsymbol{e}_t^u~(1\leq t\leq 9)$ for each 
item in the input sequence $u$, resulting in a $9\times d$ matrix; Another one is the $d$-dimensional embeddings of 7 relations, $[\boldsymbol{m}_{r_1}, \boldsymbol{m}_{r_2},\cdots, \boldsymbol{m}_{r_7}]^{\top}$. The first component $9\times d$ matrix is then input to \emph{Long-term Dependency Modeling} layer that outputs a sequence of hidden vectors, i.e., $[\boldsymbol{z}_1^u, \boldsymbol{z}_2^u,\cdots, \boldsymbol{z}_t^u,\cdots,\boldsymbol{z}_9^u]^{\top}$.
Then, we define the most recent $n$ relations latent vectors (i.e., $[\boldsymbol{z}_7^u, \boldsymbol{z}_8^u, \boldsymbol{z}_9^u]^{\top}$) as the matrix $\mathbf{H}^u\in\mathbb{R}^{n\times d}$.
$\mathbf{H}^u$ is then input into the \emph{Short-term Dependency Modeling} (individual-level) layer. We can obtain the final contextual embedding $\boldsymbol{v}\in \mathbb{R}^d$ at individual-level and $\boldsymbol{g}_L\in \mathbb{R}^d$ at union-level.
Finally, our \emph{MDM} method outputs the concentration of $\boldsymbol{v}$, $\boldsymbol{g}_L$ and  $[\boldsymbol{m}_{r_1}, \boldsymbol{m}_{r_2},\cdots, \boldsymbol{m}_{r_7}]^{\top}$ as the learned embedding features for user $u$.
This embedding can be further input into traditional classification methods to detect whether user $u$ is spammer or not.

\section{Experimental Evaluation}
\label{sec:experiment}
To evaluate the effectiveness of the proposed \emph{Multi-level Dependency Model} (\emph{MDM}),
experiments were conducted on a large real-world dataset from the website \emph{www.tagged.com}.
Comparisons were made against several state-of-art methods for spammer detection on multi-relational social networks, including graph-based and sequence-based methods.
Our algorithm was implemented in TensorFlow and experiments were conducted on a computer with $28$ CPU cores and $256$GB memory. 

\subsection{Experimental Setup}
\subsubsection{Dataset}
The dataset $\footnote{The dataset we used in this paper is published with the paper ``collective spammer detection in evolving multi-relational social networks'' published on SIGKDD2015. It can be found here: $https://linqs-data.soe.ucsc.edu/public/social_spammer/?C=S;O=A$}$ used in this experiment was from \emph{www.tagged.com},
which is a website for people to meet and socialize with new friends.
The dataset contains $7$ types of directed relations,
as shown in Table~\ref{tab:7relations}.
\begin{table}[t!]  %[h]
	\centering \scriptsize
	\caption{7 relations in the \emph{Tagged.com} dataset}
	\label{tab:7relations}
	\scalebox{1.25}{	
	\begin{tabular}{c|c}
		\toprule
		Relation ID  &  Relation Name \\
		\midrule
		$r_1$ &  Give a Gift\\  
		$r_2$ &  Add Friend\\  
		$r_3$ &  View Profile\\  
		$r_4$ &  Message\\  
		$r_5$ &  Pet Game\\  
		$r_6$ &  Meet-Me Game\\  
		$r_7$ &  Report Abuse\\  
		\bottomrule
	\end{tabular}
	}
\end{table}
The ground truth label is provided by domain experts to mark each user as normal user or spammer.
Specifically, the domain experts manually reviewed all users receiving a high number of ``abuse reports'' and terminated their accounts once confirmed.
This dataset is a benchmark data for sophisticated spammers identification on multi-relational social networks. For fair comparison, we use the same extraction process as in~\cite{fakhraei2015collective}.
The data is stored as quad-tuples:
$\langle \text{timestamp}, u_i^{\text{src}}, u_j^{\text{dest}}, r_m \rangle$,
where user $u_i^{\text{src}}$ performs relation $r_m$
on user $u_j^{\text{dest}}$.
We extracted all relations of a day,
resulted in a dataset containing $85$M interactions among $4$M users,
i.e., average length of one user's relational sequence is $21$.
Out of $4$M users,
$182$K of them are labeled as spammers,
i.e., $4.45\%$.
Statistics of the dataset is shown in Table~\ref{tab:dataset}.
\begin{table}[t!]  %[h]
	\centering \scriptsize
	\caption{Statistics of \emph{Tagged.com} dataset}
	\label{tab:dataset}
	\scalebox{1.25}{	
	\begin{tabular}{l|r}
		\toprule
		Dataset  &  \emph{Tagged.com} \\
		\midrule
		\#user &  $4,111,179$\\  
		\#spammer &  $182,939$\\  
		\#normal user &  $3,928,240$\\  
		\#interactions &  $85,470,637$\\  
		AVG length of relational sequence &  $21$\\  
		\bottomrule
	\end{tabular}
	}
\end{table}

\subsubsection{Evaluation Metrics}
Since the ground-truth label of each user is provided by the dataset,
we adopt three well-known metrics including Precision (P), Recall (R) and F-measure (F) for evaluation.
We have
\begin{equation}
\label{equ:eva}
R=\frac{TP}{TP+FN},~P=\frac{TP}{TP+FP},~F=\frac{2P\cdot R}{P+R},
\end{equation}
where $TP$ is the number of spammers that have been identified correctly, 
on contrast, 
$FP$ is the number of spammers that have been mis-identified,
and $FN$ is the number of spammers that are not identified by the model.
Depending on the application scenario,
a trade-off can be made on these metrics.
Precision and recall are contradictory metrics. Higher recall indicates that more spammers are detected.
Meanwhile, as higher recall takes more users as spammers, and it may result in low precision.
Higher precision may lead to low recall, as higher precision represents for higher confidence on detected spammers thus more spammers are missed.
F-measure is a measure of trade-off between precision and recall, which is denoted as a weighted average of the precision and recall.
Here our focus is mainly on evaluating the quality of features extracted from multi-relational data, rather than comparing the classification algorithm performance. Thus, we select two most representative supervised models, namely Logistic Regression~(\emph{LR})~\cite{agresti2003categorical} and XGBoost~(\emph{XGB})~\cite{chen2016xgboost} to classify spammers. To avoid overfitting issue, we adopt From 10-fold cross-validation for selecting the optimal parameters for Logistic Regression and XGBoost. For Logistic Regression, we specify $l_2$ norm penalty with the default strength $C=1$. We also set tolerance as $0.0001$ for stopping criteria and the maximum number of iterations as 50. We use XGBoost to implement the tree-based components of all methods, where the number of trees is 200 and the maximum depth of trees is 5.

\subsubsection{Baselines}
Several state-of-the-art graph-based and sequence-based methods are chosen as the baselines.
%, including 	
%	\emph{k-core}~\cite{alvarez2006large},
%	\emph{Graph Coloring}~\cite{jensen2011graph},
%	\emph{Page Rank}~\cite{page1999pagerank},
%	\emph{Weakly Connected Components}~\cite{pemmaraju2003computational},
%	\emph{Degree}~\cite{fakhraei2015collective},
%	\emph{Triangle Count}~\cite{schank2007algorithmic} 
%	and	\emph{Sequential k-gram Features}~\cite{fakhraei2015collective}.
Specifically,
graph-based features are extracted by converting relations
into a directed graph $\mathcal{G}$,
where the vertices $\mathcal{V}$ represent the users
and the edges $\mathcal{E}$ represent the relations user performed.
In \emph{Tagged.com} dataset, there are $7$ types of relations,
one graph is generated for each of them:
$\{\mathcal{G}_1, \ldots, \mathcal{G}_7\}$.
Then, for each graph we use \emph{Graphlab Create}$\footnote{https://turi.com/}$ to extract graph-based features, including \emph{Triangle Count}~\cite{schank2007algorithmic},
\emph{k-core}~\cite{alvarez2006large}, 
\emph{Graph Coloring}~\cite{jensen2011graph},
\emph{Page Rank}~\cite{page1999pagerank},
\emph{Degree}~\cite{fakhraei2015collective},
and \emph{Weakly Connected Components}~\cite{pemmaraju2003computational}.
This converts a directed graph into either a numerical or categorical feature matrix for each kind of relation.
That thus totally generates $7\times 8$ graph-based features. The graph-based feature can be viewed as a 56-dimensional vector.

Sequential $k$-gram aims to construct the sequence by the short sequence segment of $k$ consecutive actions. The sequence can be represented as a vector of the frequencies of the $k$-grams. To keep the feature space computationally manageable, the baseline method~\cite{fakhraei2015collective} sets $k=2$, e.g., sequence 1-1 or 2-1.  That means we have 49 types of sequences for the 7 relations, which indicates the dimension of the sequence is 49. For a specific user with behaviour of 1-1-2-3, the corresponding 49 dimensional sequence vector is $[1,1,0,0,0,0,0,0,0,1,0,\cdots,0]$.

\subsection{Experimental Results}
The embedding size $d$ in our \emph{MDM} is chosen from $\{8, 16, 32\}$, where $d=32$ produces the best results on all three metrics.
We use the most recent $n$ relations for short-term dependency modeling,
where $n$ is chosen from $\{2,4,6,8\}$ with $n=6$ producing the best results on F-measure and precision.
We also try different number of hidden layers of $ResNet^R$ and $ResNet^E$,
from $\{2,4\}$, 
as we find that 4 layers 
are enough to ensure competitive results for both $ResNet^R$ and $ResNet^E$.

After getting all the features from baseline methods and \emph{MDM},
we split train and test dataset with $10$ different random seeds
for evaluation on \emph{LR} and \emph{XGB} classifiers.
First, we compare our \emph{MDM} with them separately.
Then, we combine the baseline methods together to show the effectiveness
of our proposed model.

\subsubsection{Overall Comparison with Baselines}
Table~\ref{tab:eva} shows the comparison performance of our \emph{MDM} and baselines.
We can find that the higher recall rate happens along with the lower precision. That means normal users may be falsely identified as spammers to guarantee more spammers are detected. In this case, recall rate and precision are not sufficient to verify the effectiveness of our method. We further introduce F-measure to evaluate our performance by computing the harmonic mean of the precision and recall.
As can be seen, \emph{MDM} has shown a significant performance advantage over baseline methods on F-measure both with \emph{LR} and \emph{XGB},
which means we can catch the spammer more accurately with the least harm to normal users.
Encouragingly, the precisions of \emph{MDM} consistently are the highest ones with the best performing parameters ($d=32,n=6,k=4$),
giving the proof that the proposed features can reveal the most of
spammers with a little loss in recalls.
In terms of recall, although sequential $k$-gram features enjoy the highest position,
they show the worst performance on precision as the price,
which means they treat more users as spammers and greatly affect the normal users.
\begin{table*}[h]
	\centering
	\caption{Performance comparison with baselines (the best result of each metric is bold)}
	\label{tab:eva}
	\scalebox{0.75}{
		\begin{tabular}{@{}lcccccccccc@{}}
			\toprule
			& \multicolumn{3}{c}{Logistic Regression~(\emph{LR})} & \multicolumn{3}{c}{XGBoost~(\emph{XGB})} \\ \cmidrule(l){2-4}\cmidrule(l){5-7}
			Methods & Precision & Recall & F-measure & Precision & Recall & F-measure\\ \midrule
			Graph-based~\cite{brophy2017collective} & 0.5576 & 0.6937 & 0.6182 & 0.6378 & 0.6712 & 0.6541 \\
			\cmidrule(l){2-4}\cmidrule(l){5-7}
			Sequential $k$-gram~\cite{fakhraei2015collective} & 0.5217 & \textbf{0.8620} & 0.6500 & 0.5268 & \textbf{0.9221} & 0.6705 \\
			\cmidrule(l){2-4}\cmidrule(l){5-7}
			Graph-based+Sequential $k$-gram & 0.6116 & 0.8600 & 0.7148 & 0.6253 & 0.9127 & 0.7421 \\
			\cmidrule(l){2-4}\cmidrule(l){5-7}
			\cmidrule(l){2-4}\cmidrule(l){5-7}
			\textbf{MDM} & \textbf{0.6909} & 0.8243 & \textbf{0.7516} & \textbf{0.7385} & 0.8154 & \textbf{0.7750} \\
			\bottomrule
		\end{tabular}
	}
\end{table*}

\subsubsection{Effect of Parameters within MDM}
\begin{figure*}[htbp]
	\centering
	\subfigure[Precision]{
		\includegraphics[width=0.3\linewidth]{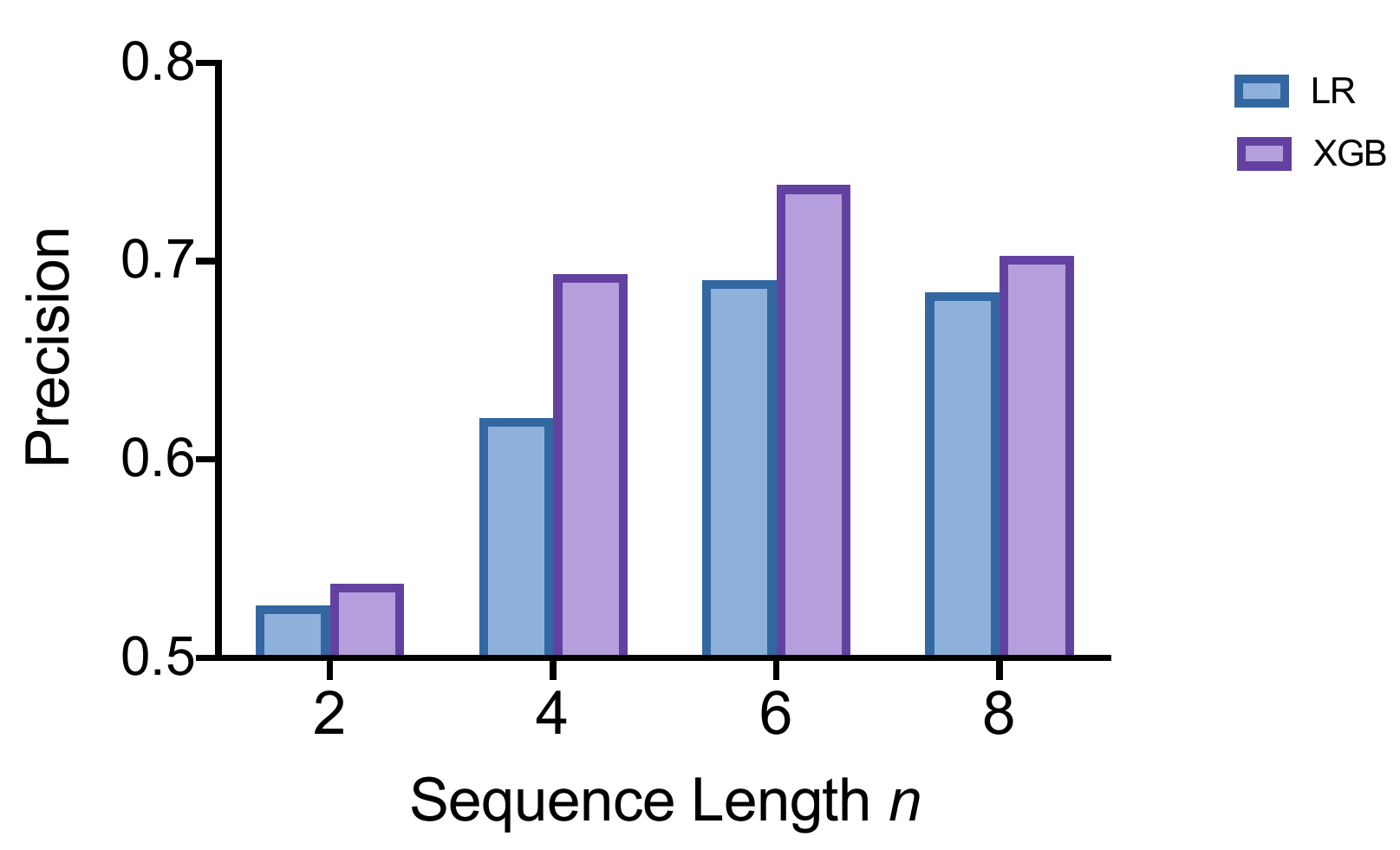}
		\label{fig:P_n}}
	\subfigure[Recall]{
		\includegraphics[width=0.3\linewidth]{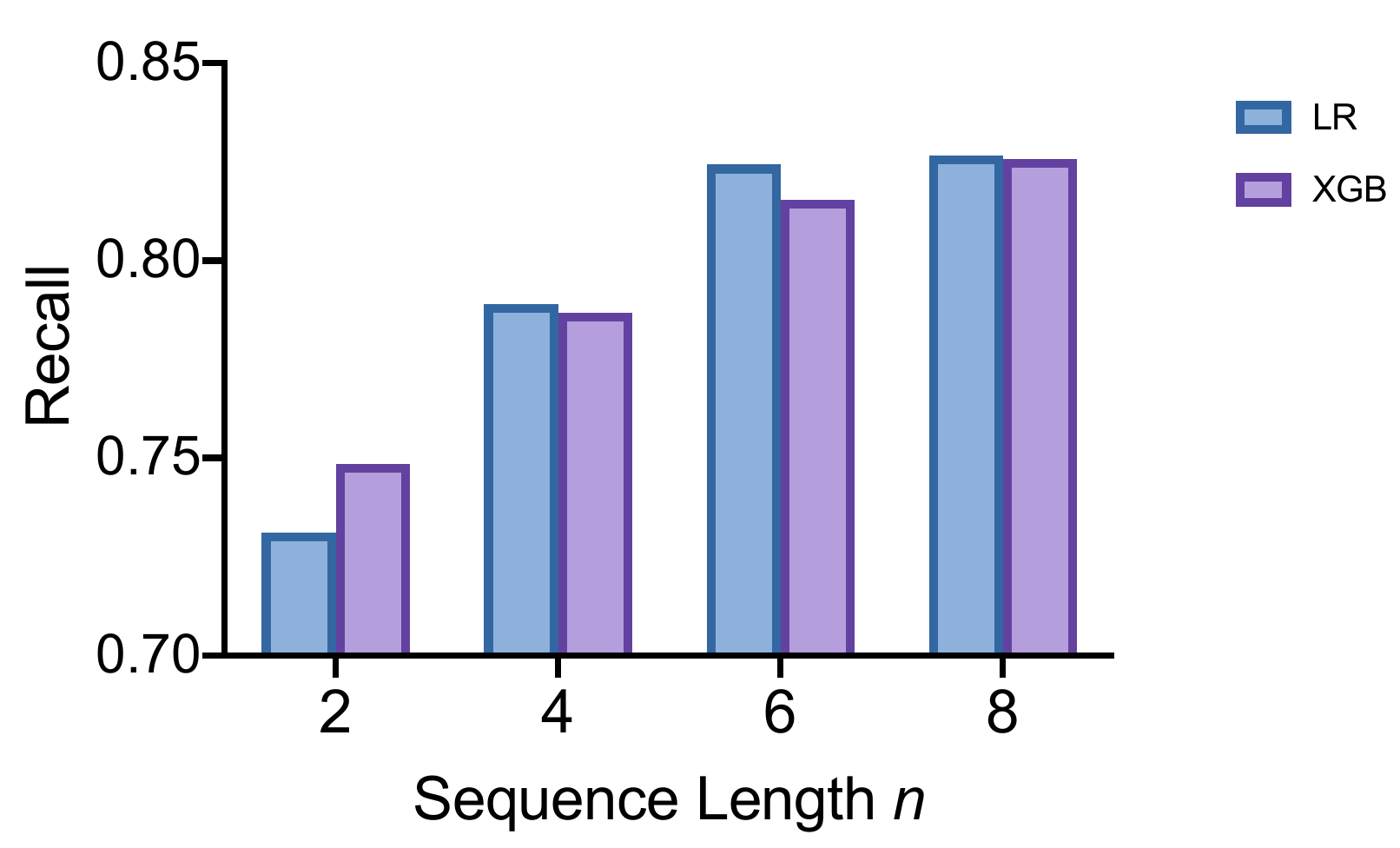}
		\label{fig:R_n}}
	\subfigure[F-measure]{
		\includegraphics[width=0.3\linewidth]{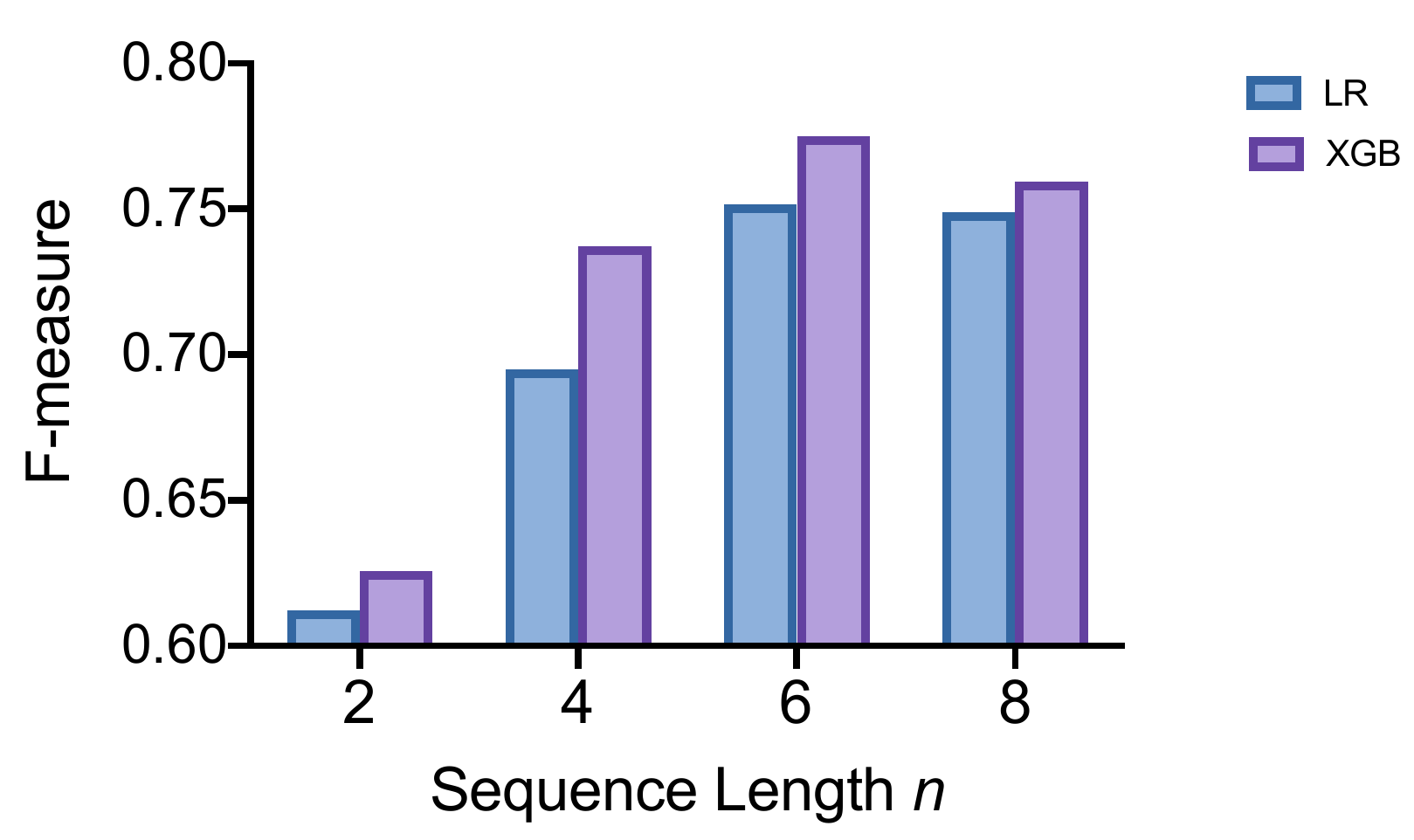}
		\label{fig:F_n}}
\caption{Performances of MDM under different sequence lengths $n$.}
\label{fig:parameter_n}
\end{figure*}

We further evaluate the performance of the~\emph{MDM} with respect to the parameter settings.
First, we vary the sequences length in short-term information modeling. The comparison is set on different $n$ chosen from $\{2,4,6,8\}$. Fig.~\ref{fig:parameter_n} shows the comparison results of different setups. The results show that when other parameters are set equal, $n=6$ promotes the best performance.
One presumable assumption is that 6 steps of behaviours can better summarize a user's intention in \emph{Tagged.com}.

\begin{figure*}[htbp]
	\centering
	\subfigure[Precision]{
		\includegraphics[width=0.3\linewidth]{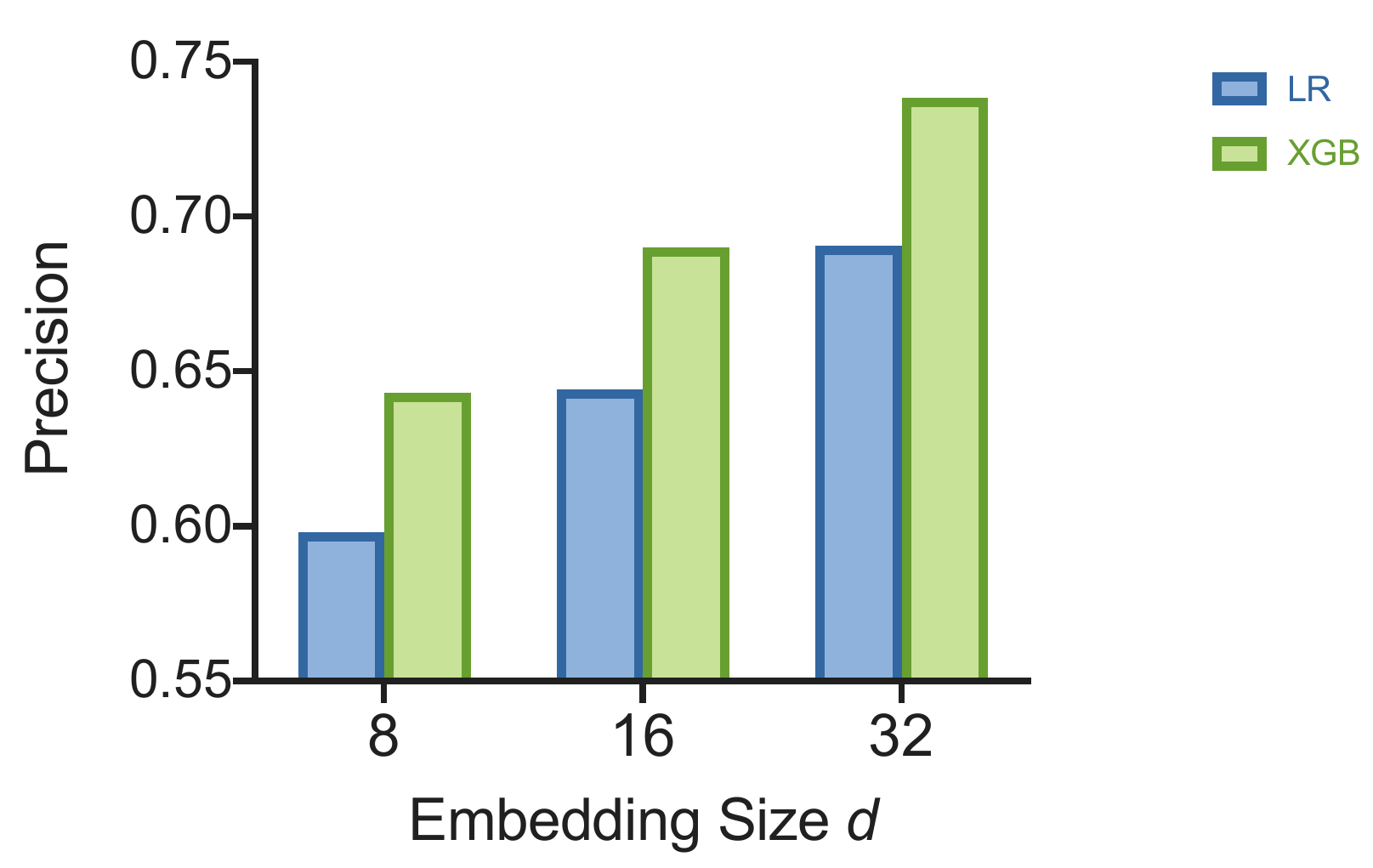}
		\label{fig:P_d}}
	\subfigure[Recall]{
		\includegraphics[width=0.3\linewidth]{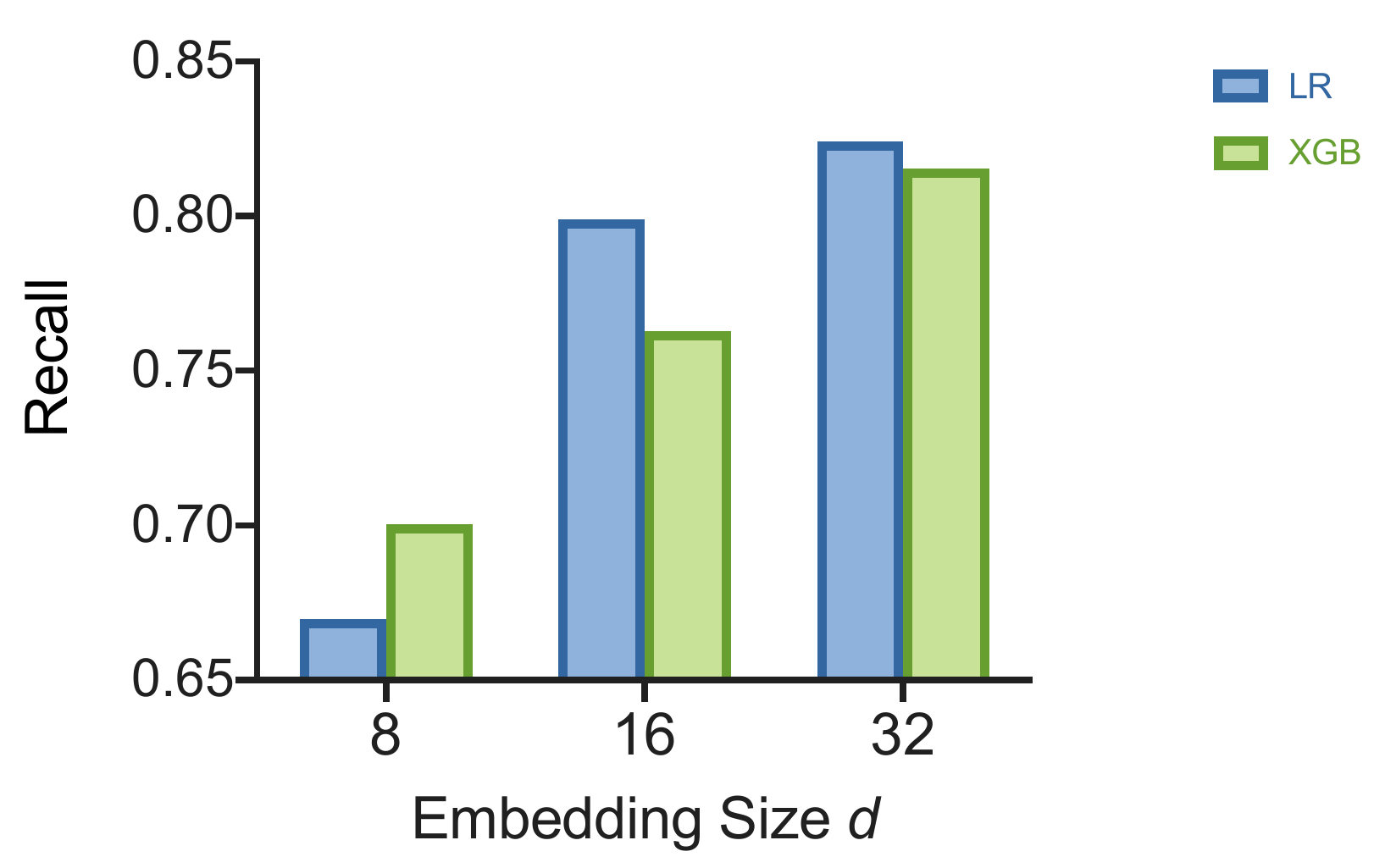}
		\label{fig:R_d}}
	\subfigure[F-measure]{
		\includegraphics[width=0.3\linewidth]{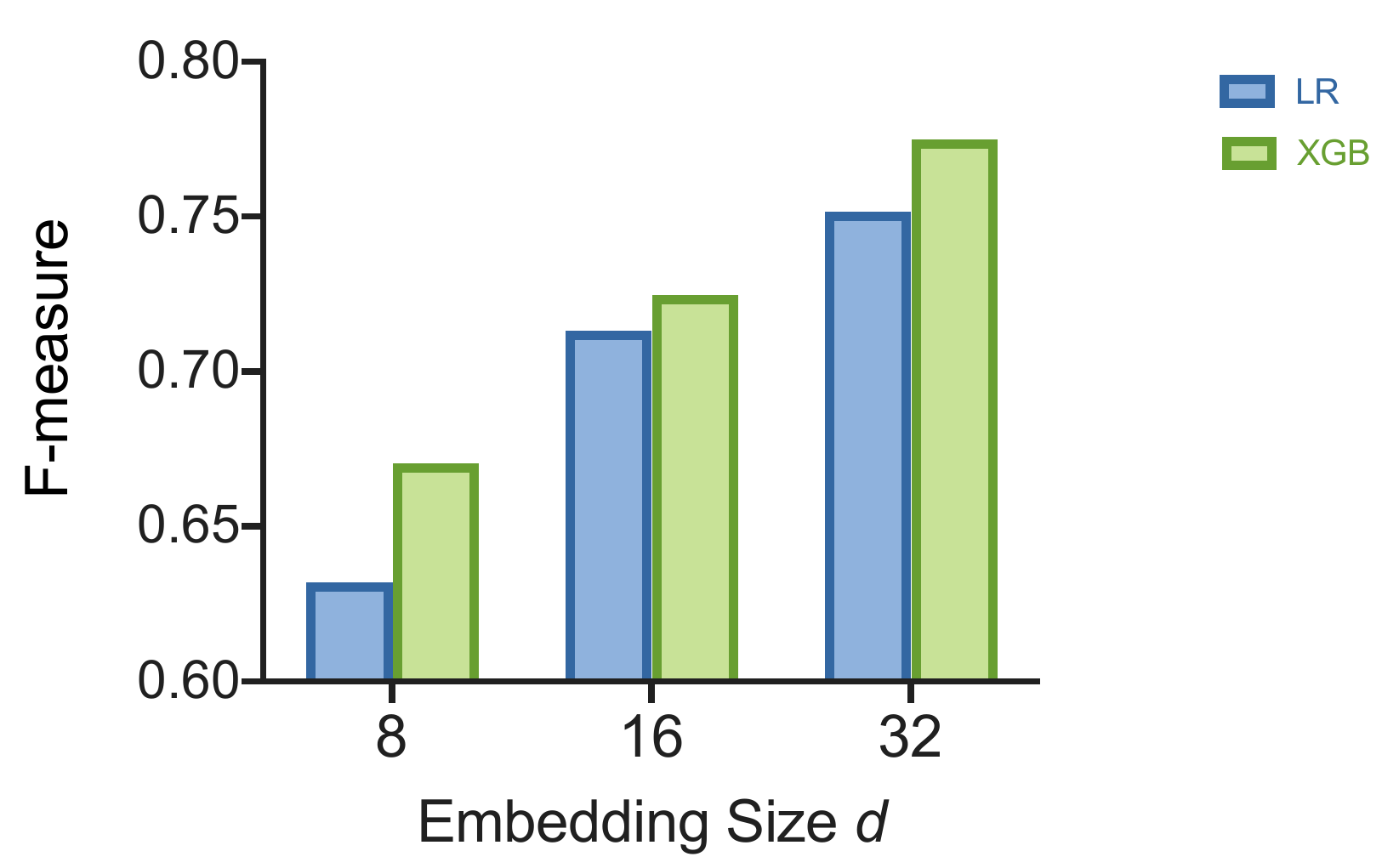}
		\label{fig:F_d}}
\caption{Performances of MDM under different embedding sizes $d$.}
\label{fig:parameter_d}
\end{figure*}

Then we analyze \emph{MDM}'s performance by varying the embedding size $d$ from $\{8,16,32\}$.
Fig.~\ref{fig:parameter_d} shows the performance of each size of embedding features on precision, recall and F-measure separately.
Obviously, the three metrics' rates increase with the raise of dimension, giving the sign that more spammers will be disclosed when increasing the embedding dimension of our~\emph{MDM} and more accurate it will be.

In general, it shows that the most effective performance has been achieved on 32 embedding features. 
Limited by the computing space, we only carry the embedding size to 32.
Nevertheless, the number of embedding size depends on the dataset.
One recommendation is that the number of embedding features should be increased alongside the number of types of relations,
because more type of relations implies more complex interactions.

\subsubsection{Components Influence of MDM}
\emph{MDM} contains three components as indicated in Fig.~\ref{fig:framework}, 
i.e. \emph{User-relation Representation}, 
\emph{Long-term Dependency Modeling}, 
and \emph{Short-term Dependency modeling}, 
where the last component is made up of individual-level and union-level.
In order to analyze the impact of the different components to the overall detection performance, we set different combinations of components for evaluation. The comparison results are shown in Table~\ref{tab:component}. 

It can be seen from the table that although \emph{User-relation Representation} obtains the highest recall on both \emph{LR} and \emph{XGB},
its precision and F-measure score are the lowest.
After adding the \emph{Long-term Dependency Modeling} layer,
the precision increases with a little drop in recall and the overall F-measure score rise on both \emph{LR} and \emph{XGB}.
That is to say,
\emph{Long-term Dependency Modeling} layer can help lifting the social spammer detection performance as we estimated.

After we take \emph{Short-term Dependency modeling (individual-level)} into consideration,
we can see from the table that both precision and F-measure have been greatly improved, indicating that the short-term dependency within the relational sequence largely boosts with the users' hidden sequential information modeling. 
Afterwards, the proposed \emph{MDM}, 
consisting of all three layers, gets the best performance on precision and F-measure with the best performing parameters ($d=32,n=6,k=4$). In other words, with \emph{MDM} we can detect more spammers correctly and without harming normal users.

\begin{table*}[h]
	\centering
	\caption{Performance comparison on different components in MDM (+ represents adding a layer to the last row, and the best result of each metric is bold)}
	\label{tab:component}
	\scalebox{0.71}{
		\begin{tabular}{@{}lcccccccccc@{}}
			\toprule
			& \multicolumn{3}{c}{Logistic RegressionD~(\emph{LR})} & \multicolumn{3}{c}{XGBoost~(\emph{XGB})} \\ \cmidrule(l){2-4}\cmidrule(l){5-7}
			Components & Precision & Recall & F-measure & Precision & Recall & F-measure\\ \midrule
			User-relation Representation & 0.5399 & \textbf{0.8496} & 0.6602 & 0.5778 & \textbf{0.8722} & 0.6951 \\
			\cmidrule(l){2-4}\cmidrule(l){5-7}
			+ Long-term & 0.5687 & 0.8467 & 0.6804 & 0.5937 & 0.8718 & 0.7064 \\
			\cmidrule(l){2-4}\cmidrule(l){5-7}
			+ Individual-level & 0.6314 & 0.8477 & 0.7237 & 0.6659 & 0.8523 & 0.7477 \\
			\cmidrule(l){2-4}\cmidrule(l){5-7}
			\cmidrule(l){2-4}\cmidrule(l){5-7}
			\textbf{MDM} & \textbf{0.6909} & 0.8243 & \textbf{0.7516} & \textbf{0.7385} & 0.8154 & \textbf{0.7750} \\
			\bottomrule
		\end{tabular}
	}
\end{table*}

\subsection{Discussion}
We have studied the sequences of behaviours in the multi-relational social network (i.e.,\emph{Tagged.com}) to detect unknown spammers. 
From our experiments, we found some interesting spamming behaviour. Our results indicate that users with the sequences $\langle 5,5,5,5,5,5 \rangle$, $\langle 5,5,5,5,5,4 \rangle$, $\langle 4,4,3,5,4,4 \rangle$ are easily detected as spammers in our proposed \emph{MDM}. However, the dataset collected from \emph{Tagged.com} does not connect the relation names with the specific number. Namely, from the dataset, we do not know which number corresponds to which relation. To fully understand the behaviour behind this sequence, we use statistic analysis technology to infer what relationship does the number represent for.
For example the sequence $\langle 5,5,5,5,5,5 \rangle$. We infer that $5$ is a ``Pet Game''. $\langle 5,5,5,5,5,5 \rangle$ means the user has been always playing this ``Pet Game''. The users have such a behaviour sequence is recognized as the spammer. Because \emph{Tagged.com} has a reward mechanism for ``Pet Game''. In order to be seen/contacted by more users, the spammers always gaining more reward by playing ``Pet Game'' will appear on the celebrity list.
In addition to such special sequences, we find that a user that repeats a single relation and occasionally transforms one or two relations to hide its behaviour is more likely to be a spammer.

\section{Conclusion}
\label{sec:conclusion}
In this work, we propose a novel \emph{Multi-level Dependency Model (MDM)} to fully learn the deeper complementary information underlying users' relational sequences. 
\emph{MDM} exploit user's behaviours in terms of the \emph{long-term} and \emph{short-term} dependencies. In particular, the \emph{short-term} dependency can be sufficiently exploited in terms of both individual-level and union-level.
Therefore, \emph{MDM} is capable of exposing the deep information underlying the relational sequences so as to improve the accuracy of identifying abnormal behaviours.
We conduct extensive experiments on real-world dataset from \emph{Tagged.com}, which verifies that \emph{MDM} significantly outperforms other baselines.
The main limitation of this work, due to privacy concerns, it is focused on relations (i.e., interactions between users) only without considering the free text information. In the future, it is worth exploring the possibility of incorporating text information into the modelling process, for example, applying to a Twitter data set.
In order to generate \emph{MDM} on other multi-relational data set, e.g., Twitter data set, we need to decompose the user behaviour into several relations, then pick up all relations without duplication, and assign each of them with a unique number.

\bibliography{mybib}

\end{document}